\documentclass{article} %
\usepackage{arxiv,times}

\usepackage{amsmath,amsfonts,bm}

\def\eqref#1{equation~\ref{#1}}

\def\1{\bm{1}}

\DeclareMathAlphabet{\mathsfit}{\encodingdefault}{\sfdefault}{m}{sl}
\SetMathAlphabet{\mathsfit}{bold}{\encodingdefault}{\sfdefault}{bx}{n}

\usepackage[normalem]{ulem}

\usepackage{graphicx}
\usepackage{subcaption}
\usepackage{wrapfig}
\usepackage{url}
\usepackage{booktabs}       %
\usepackage{amsfonts}       %
\usepackage{nicefrac}       %
\usepackage{microtype}      %
\usepackage{xcolor}         %
\usepackage{listings,enumitem}
\usepackage{subcaption}

\definecolor{myblue}{rgb}{0,0,0.5}
\usepackage[colorlinks,citecolor=myblue,urlcolor=myblue,linkcolor=myblue]{hyperref}
\usepackage[capitalise,noabbrev]{cleveref}
\usepackage[hide=false,setmargin=true,marginparwidth=1.2in]{marginalia}

\usepackage{wrapfig}
\usepackage{makecell}
\usepackage{multirow}

\usepackage{amssymb}%
\usepackage{pifont}%
\newcommand{\cmark}{\ding{51}}%
\newcommand{\xmark}{\ding{55}}%

\title{The Cost of Down-Scaling Language Models: \\ Fact Recall Deteriorates before In-Context Learning}

\iclrfinalcopy

\author{Tian Jin\thanks{These authors contribute equally to this work.} \\
MIT CSAIL \\
\texttt{tianjin@csail.mit.edu} \\
\And
Nolan Clement\footnotemark[1] \\
MIT \\
\texttt{nolangc@mit.edu} \\
\And
Xin Dong\footnotemark[1] \\
Harvard University \\
\texttt{xindong@g.harvard.edu} \\
\And
Vaishnavh Nagarajan \\
Google Research \\
\texttt{vaishnavh@google.com} \\
\And
Michael Carbin \\
MIT CSAIL \\
\texttt{mcarbin@csail.mit.edu} \\
\And
Jonathan Ragan-Kelley \\
MIT CSAIL \\
\texttt{jrk@mit.edu} \\
\And
Gintare Karolina Dziugaite \\
Google DeepMind \\
\texttt{gkdz@google.com}
}

\begin{document}

\maketitle

\begin{abstract}

How does scaling the number of parameters in large language models (LLMs) affect their core capabilities? 
We study two natural scaling techniques — weight pruning and simply training a smaller or larger model, which we refer to as dense scaling — and their effects on two core capabilities of LLMs: (a) recalling facts presented during pre-training and (b) processing information presented in-context during inference. By curating a suite of tasks that help disentangle these two capabilities,
we find a striking difference in how these two abilities evolve due to scaling. 
Reducing the model size by more than 30\% (via either scaling approach) significantly decreases the ability to recall facts seen in pre-training. 
Yet, a 60--70\% reduction largely preserves the various ways  the model can process in-context information, ranging from retrieving answers from a long context to learning parameterized functions from in-context exemplars. 
The fact that both dense scaling and weight pruning
exhibit this behavior suggests that scaling model size has an inherently disparate effect on fact recall and in-context learning.

\end{abstract}

\section{Introduction}
Scaling up the size of LLMs is known to yield
impressive performance gains on various tasks \citep{DBLP:journals/corr/abs-2001-08361,hoffmann2022training, brown2020language, wei2022emergent}.  On the other hand, to deploy language models sustainably, it is also critical to scale them \textit{down} while preserving their end utility.
Naturally, scaling, in both directions, has gained interest in a recent wave of research on language models
(\citet{DBLP:journals/corr/abs-2001-08361,hoffmann2022training, frantar2023sparsegpt,jiang2022pruning,kurtic2023ziplm,santacroce2023matters}). %
Much of this work evaluates size--performance tradeoffs of scaling through aggregate performance metrics such as 
perplexity or downstream accuracy on existing benchmarks.

However, we argue that there must exist subtle but important effects of scaling that cannot be captured by standard metrics alone.
Indeed, work on image classification already hints at this.
Pruning image models, for example, can introduce biases \citep{hooker19} or disproportionate effects on certain subsets of the data \citep{NEURIPS2022_f7ede941} --- these are effects that simply do not reflect in the overall accuracy of the model. 
Taking inspiration from this, our goal is to identify the subtler effects of scaling (up or down) LLMs in terms of the various \textit{capabilities} that underpin their success in practice.

\begin{figure}
  \centering
  \includegraphics[width=.98\textwidth]{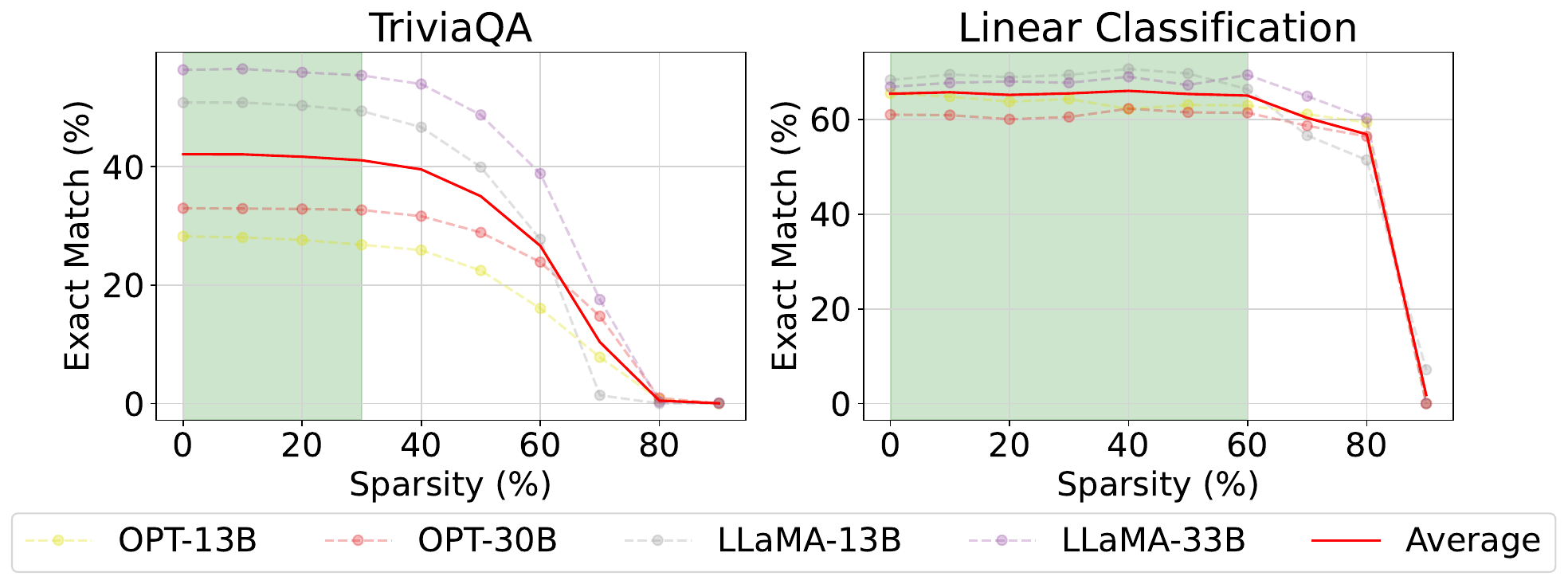}
  \caption{Pruning to moderate sparsity ($>30\%$ sparse) harms fact recall while in-context learning withstands even aggressive pruning (60\% sparse).
    We plot the accuracy versus sparsity for the OPT-13B, OPT-30B, LLaMA-13B, and LLaMA-33B models on TriviaQA dataset for fact recall evaluation (left) and 
    linear classification dataset for ICL evaluation (right).
    We plot their average performance in red.
    We color the range of sparsity where accuracy drop is within (relative) 5\% of the dense model.
    Accepting (relative) 5\% accuracy drop w.r.t. the dense model, the maximum sparsity on fact recall task is 30\%, whereas the maximum sparsity on ICL task is 60\%. }
  \label{fig:intro}
\end{figure}

\textbf{Our approach.} 
In this work, we study the effects of scaling the number of parameters in an LLM on two fundamentally dichotomous capabilities as put forth by \citet{chan2022data,chan22differently}:
the ability to
 process information stored in  weights (such as facts seen during pre-training) and the ability to process information stored in context (such as hints given in the context). 
It is, however, challenging to isolate how well a model has learned these abilities by simply measuring performance on an arbitrary downstream task --- after all, most tasks require {both} abilities to varying extents to succeed.
Instead, we carefully curate a set of downstream tasks
that help tease apart the performance of any given model on the two abilities  (as explained shortly).
We then evaluate model performance in terms of these two abilities under two fundamentally different types of scaling: \textit{pruning} and \textit{dense scaling}. 
For the former, we consider recent scalable pruning methods that prune and update the remaining weights in one-shot \citep{frantar2023sparsegpt, sun2023simple}; for the latter, we use (separately-trained) dense models with increased/reduced width and depth.

Our curated suite of benchmarks involve four classes of tasks, 
defining a \textit{gradation} of reliance on  information stored in-weights and in-context. 
The first three are Q\&A tasks: 
(1) a {closed book} Q\&A that necessarily requires {\em fact recall} from pre-training for which the model needs to process information from the weights,
(2) an {open book Q\&A} for which it suffices to {\em copy information} from the context and 
(3) an {overriding Q\&A}, where the information provided in-context \textit{overrides} facts seen during pre-training  \citep{li22controllable,longpre21entity}. 
Finally, to evaluate more sophisticated context-processing abilities, we consider testing the model on its ability to (4) 
 learn parameterized functions (e.g., linear classifiers) using input--output pairs presented in-context.
Overall, we consider open-book, overriding Q\&A and learning tasks as tasks that require increasing levels of in-context learning (ICL) abilities (See \Cref{subsec:prelim-eval} for more discussion). %
We summarize these tasks in Table~\ref{tab:all_datasets}.

\textbf{Fact recall results.}
Prior pruning literature demonstrates significant down-scaling without noticeable change to \textit{accuracy}: pioneering work \citep{dong2017learning} removes $\sim 45\%$ of the weights in a ResNet50 model without noticeable accuracy degradation.
\footnote{Please refer to Section 4.1, Figure 2.(a) of \citet{dong2017learning}.
Notably, \citet{dong2017learning} obtained these results similarly by pruning without retraining.}
However, when examining the effect of pruning on the ability to recall facts, we find a much different story:
removing more than 30\% of weights leads to significant ($>5\%$, relative) accuracy degradation on fact recall related tasks (\Cref{fig:intro}, left).
Fact recall shows the same striking sensitivity to dense scaling.

\textbf{In-context learning results.}
The seminal work of \citet{NEURIPS2020_1457c0d6} hypothesized that scaling up the number of parameters strongly benefits the ICL ability of LLMs.%
\footnote{``Since in-context learning involves absorbing many skills and tasks within the parameters of the model, it is plausible that in-context learning abilities might show similarly strong gains with scale." \citep{NEURIPS2020_1457c0d6}} 
However, our evaluations on the suite of tasks that require a gradation of ICL abilities (with little to no reliance on fact recall) demonstrate the opposite: substantial down-scaling still preserves LLM's ICL ability.
Specifically, even after pruning 60\% of weights, the relative accuracy decrease on ICL tasks is less than 5\% (\Cref{fig:intro}, right).
ICL shows a similar \textit{insensitivity} to dense scaling.

\textbf{Implications.}
Our work reveals the disparate effects of scaling on fact recall and in-context learning (ICL), which notably holds for two substantially different types of scaling: pruning and dense scaling.  This motivates several lines of future work:

\emph{Improve inference efficiency.}
Our work reveals that scaling down model size alone has little impact on tasks demanding processing information in the LLM's context.
Practitioners may thus use our findings to identify scenarios where decisions could be routed to a smaller model instead of a larger one without hurting task performance \citep{chen2023frugalgpt, dohan2022language}.

\emph{Improve LLM systems.} 
Our work shows the utility of feeding  external supportive information, like documents from similarity search \citep{pmlr-v162-borgeaud22a, 10.5555/3524938.3525306}, in question--answering tasks to improve the compute cost and task accuracy trade-off for down-scaled models.

\emph{Improve LLM Interpretability.}
We find a remarkably small set of weights responsible for ICL.
This underscores the potential of pruning as a tool for isolating neurons responsible for LLM capabilities.

\textbf{Contributions.}
\begin{enumerate}[leftmargin=1.5em]
  \item We curate a set of benchmarks for assessing the disparate effects of down-scaling on core capabilities of LLMs, focusing on fact recall and ICL.
  \item Using these benchmarks, we evaluate two  scalable weight pruning algorithms for LLMs. 
  We investigate an extensive set of models comprising six base LLMs, with sizes reaching up to 33 billion parameters.
  We find that:
  \begin{itemize}
      \item even moderate levels of pruning ($> 30\%$) hurt fact recall. However, when the evidence required to solve a question-answering task is provided in context, the model's ability to correctly answer questions survives to higher levels of sparsity;
      \item in contrast, in-context learning withstands aggressive pruning (up to 60--70\%).
  \end{itemize}

  \item Similarly for dense scaling, we find  the same disparate patterns as above for fact recall and ICL abilities relative to model size.  This underscores that the disparate effects on ICL and fact recall are not exclusive to pruning but are broader characteristics of scaling in general. 
  
\end{enumerate}

\begin{table}[t]
    \small
    \caption{Tasks employed in our experiments.}
    \label{tab:all_datasets}
    \centering
    \resizebox{\columnwidth}{!}{%
    \begin{tabular}{clccccc}
    \toprule
    \textbf{Task Type} & \textbf{Task} & \makecell{\textbf{Context} \\ \textbf{Type}} & \makecell{\textbf{Recalling Facts}\\ \textbf{from Pre-training}} & \makecell{\textbf{Retrieving Facts}\\ \textbf{from Contexts}} &  \makecell{\textbf{Learning Patterns}\\ \textbf{from Contexts}} & \makecell{\textbf{Considered}\\ \textbf{ICL}} \\
    \midrule
    \multirow{5}*{Q\&A} & WebQuestions & Empty & \cmark & \xmark & - & \xmark \\ 
    & NaturalQuestions & Factual & \cmark & \cmark & - & \cmark \\
    & TriviaQA & Empty  & \cmark & \xmark & - & \xmark \\
    & TriviaQA (Filtered) & Factual  & \cmark & \cmark & - & \cmark \\
    & DisentQA & Synthetic & \xmark & \cmark & - & \cmark \\
    \midrule
    \multirow{3}*{\makecell{Learning}} & Linear CLS & Exemplary & - & - & \cmark & \cmark  \\
    & 2-layer NN & Exemplary & - & - & \cmark & \cmark  \\
    & Decision Tree & Exemplary & - & - & \cmark & \cmark  \\
    \bottomrule
    \end{tabular}
    }
\end{table}

\section{Related Work}

\textbf{LLM scaling.}
Studies on scaling laws \citep{hoffmann2022an, kaplan2020scaling} suggest a predictable relationship between the quality of language models (i.e., perplexity) and the size of the pre-training corpus and model.
Pursuant to this, many have scaled up and discovered remarkable capabilities of language models.
\citet{brown2020language} discover that LLMs can perform in-context learning (ICL) effectively: the model learns to perform tasks based on a few examples of input--output demonstrations in the model's context.
Other studies \citep{devlin-etal-2019-bert, wei2022emergent, liang2022holistic, srivastava2022imitation, eval-harness, 10.1145/3531146.3533229, bubeck2023sparks, biderman2023emergent} benchmark LLMs across many tasks and metrics to assess the utility of scaling.
Our work differs in two ways: while prior work \citep{kaplan2020scaling} studies joint scaling of both pre-training corpus size and model size, we focus on scaling model size alone.
Furthermore, instead of measuring an arbitrary notion of task performance, we focus on the foundational capabilities of fact recall and ICL. 
These capabilities underpin the success of many real world applications of LLMs.

\textbf{In-weight versus in-context learning.}
LLMs utilize information both stored in weights and present in context \citep{chan22differently}. 
Recent research illustrates the similarities and differences between these learning approaches:
\Citet{vonoswald2022transformers} demonstrated that in-context learning (ICL) can implement an algorithm akin to gradient descent, commonly associated with in-weight learning.
\citet{akyurek2023what} revealed that ICL resembles various in-weight learning algorithms, depending on model size. 
\citet{chan2022data, chan22differently} study how 
how the properties of the data distribution disparately affect the two abilities in the model. 
Our work identifies yet another crucial difference: scaling model size has distinct impacts on in-weight learning versus in-context learning abilities.

\textbf{Neural network pruning.}
Pruning removes unimportant parameters in a model. 
The origin of pruning traces back to \citet{NIPS1989_6c9882bb, 298572}, 
with a focus on reducing the computational footprint. 
More recently, with the advent of deep learning, pruning research has seen a resurgence \citep{Renda2020Comparing,DBLP:journals/corr/HanMD15,liu2017learning, frankle2021pruning,molchanov2017pruning, NIPS2017_c5dc3e08, ma2023llmpruner,
tao-etal-2023-structured, kurtic2023ziplm, dettmers2023spqr}.
Though pruning traditionally focuses on preserving aggregate metrics such as accuracy, the versatility of LLMs calls for a different approach to assessing pruned models.
Our work begins to fill this gap, proposing to evaluate pruning's effect on fact recall and ICL.
Our investigation extends a growing line of studies on effects of pruning beyond aggregate metrics such as accuracy and perplexity. 
\citet{hooker19, DBLP:journals/corr/abs-2010-03058} show that pruning may harm under-represented categories of examples; 
\citet{MLSYS2021_2a79ea27} suggest that the pruned models are less accurate than the dense one when predicting out-of-distribution examples.
\citet{NEURIPS2022_f7ede941} demonstrate that pruning mitigates the harmful effects of noisy labels on generalization. 

Concurrent to our work, \citet{yin2023junk,
jaiswal2023compressing} remarked on the pronounced negative effect of pruning on knowledge-intensive tasks.
These works extensively evaluate tasks that require abilities of fact recall and in-context information retrieval.
Our work additionally curates a gradation of tasks that further \textit{disentangle} these abilities --- specifically via overriding QA and parameterized ICL tasks, which are not covered in these works.

\section{Preliminaries}
\label{sec:prelim}
 
\textbf{Pruning algorithms.}
We investigate pruning as one possible technique to (down-)scale LLMs. Few pruning algorithms currently scale to LLMs.
We use SparseGPT \citep{frantar2023sparsegpt} in the main text and Wanda \citep{sun2023simple} in \Cref{sec:another-pruning-algo}. 
Both are one-shot pruning algorithms that scale to LLMs and outperform magnitude pruning (i.e., pruning the smallest magnitude weights), 
without computationally intensive re-training \citep{frantar2023sparsegpt}.
SparseGPT/Wanda prune each layer of the language model by minimizing the $\ell_2$-distance between the outputs of the original dense layer and the pruned layer. 
SparseGPT/Wanda computes
these outputs based on a small training dataset.
See \citet[][Sec.~3]{frantar2023sparsegpt} for more details.
While SparseGPT update the remaining weights after weights removal, Wanda does not.

Following standard practice \citep{frantar2023sparsegpt, frankle2018the}, we only prune fully-connected layers.
Since attention and feed forward modules consist of mostly fully-connected layers, parameters in fully-connected layers account for $> 97.5\%$ parameters for all models we examine.
We do not prune embedding layers, language modeling heads and normalization layers.

\textbf{Models.}
We evaluate 6 models from 3 families: OPT \citep{zhang2022opt}, LLaMA \citep{touvron2023llama} and Pythia \citep{biderman2023pythia}.
We focus on OPT and LLaMA in our main text and present Pythia results in \Cref{sec:pythia}.
Pythia family models show consistent results as LLaMA and OPT family models.
From the OPT family, we evaluate the two largest models that fit in our hardware setup -- OPT-13B and OPT-30B, with 13 and 30 billion parameters, respectively.
From the LLaMA family, we evaluate LLaMA-13B and LLaMA-33B, with 13 and 33 billion parameters, respectively.

A notable difference between OPT and LLaMA families is the ratio of training data to model parameters.
\citet{zhang2022opt} train the OPT family of models with 180 billion tokens, yielding approximately 14 and 5.5 tokens \textit{per} parameter, respectively, for our two considered OPT models.
\citet{touvron2023llama} train the LLaMA-13B model with 1 trillion tokens (77 tokens/parameter) and the LLaMA-33B model with 1.4 trillion tokens (42 tokens/parameter).
\subsection{Evaluation}
\label{subsec:prelim-eval}
We evaluate models on two complementary abilities: the ability to call on information 
seen during pre-training (and then stored in the model's weights) and the ability to call on information presented in-context. 
Unfortunately,
it is difficult to construct tasks that strictly isolate one of these two abilities. 
On the one hand, for every conceivable natural language task, the model must rely on the semantic representations of language learned from pre-training and stored in the weights. 
On the other, every conceivable task necessitates  processing all of the context to understand the instructions in the first place. %
To help disentangle the two (overly) broad abilities, we focus on two well-defined instantiations of these abilities:  (a) the ability to recall {\em facts} from pre-training to answer questions and, complementing that (b) the ability to \textit{learn} patterns from context.   

\textbf{Fact recall.} 
We isolate (a) by evaluating the model on recalling facts from training data not provided in-context. We do this using closed-book QA benchmarks. 

\textbf{In-context learning.}
To evaluate (b), we consider a gradation of tasks that require the increasingly advanced ways of learning from in-context information.
Broadly, we consider a series of in-context learning tasks: tasks where the context contains examplars of (query, answer) pairs in some form, followed by a test query that the model has to answer.

First, as a simple ICL task, we consider an Open-book QA (\cref{sec:qatasks}) counterpart of our QA task. Here the context is augmented with supporting evidence %
that directly helps answer the question. This can be thought of as a simple ICL task where the test query (``who is the author of the novel The Eagle Has Landed?'') is simply already present as an in-context examplar within the evidence (``The author of The Eagle Has Landed is...''), along with its ground truth answer (e.g., ``...Jack Higgins"). 
Note that in addition to the aforementioned evidence--question--answer triplet, we further aid the model with an independent one-shot triplet to demonstrate how to format its answer.

In the above case, the answer provided in-context may also have been present in pre-training data; thus the model may opt to ignore the context, and recall facts from pre-training, and still succeed at the task. 
Thus, to more rigorously isolate the two mechanisms, we next evaluate on the Overriding QA task (\cref{sec:qatasks}), where the in-context evidence \textit{contradicts} a fact present in training data. 
We present a one-shot evidence--question--answer triplet, where the answer is based on the contextual evidence rather than on pre-training facts. 
Thus, to solve the task correctly, it is clear that the model must retrieve from the context rather than recall pre-training facts.  %

Next, we investigate ICL abilities more sophisticated than {\em copying} an answer present in an in-context exemplar. To this end, we are interested in tasks where the test query is not already seen as an in-context exemplar. As one such example, consider English--French translation task \citep{brown2020language} where the examplars are pairs of English and French sentences, and the test query is a new English sentence. However, such conventional instances of ICL tasks require significant background knowledge of languages and the world, not available in the context.

To ensure we disentangle the role of background knowledge from ICL, we consider tasks where the query--answer mapping is given by a \textit{parameterized} function. 
Arguably, learning this mapping,  would require advanced context-processing mechanisms that go beyond information-retrieval mechanisms required for the Open-book and Overriding QA tasks. %

\textbf{Metrics.}
As a simple point of comparison, we measure the effect of scaling on perplexity. 
However, our main metric of interest is {\em exact match accuracy} (of the answer to each task) because it is a direct measure of the model's ability to satisfactorily perform a task.

We present pruning's effects on model perplexity
in \cref{sec:pruning-perplexity-eval} and details of our hardware/software configurations in \Cref{sec:hwsw}.

\section{Effect of Pruning on Question Answering Tasks}
\label{sec:qatasks}

The ability of LLMs to answer questions underpins many applications of LLMs. %
Within a question-answering evaluation framework, we examine the effect of pruning on its ability to recall facts learnt during pre-training.
We also test a simple form of ICL capability -- extracting answers exclusively from the provided context, where answers do not exist in the pre-training corpus.
Finally, we interpolate between these two capabilities by providing supporting evidence in context when querying information is likely also available in pre-training data. 
Our results expose that, 
with contextual information in question-answering tasks, the model's ability to answer questions withstands more aggressive pruning than without contextual information.
With no supporting evidence in context, accuracy rapidly drops as sparsity increases.

\begin{figure}[t] 
    \includegraphics[width=1.\textwidth, trim={0, 11cm, 0, 0}, clip]{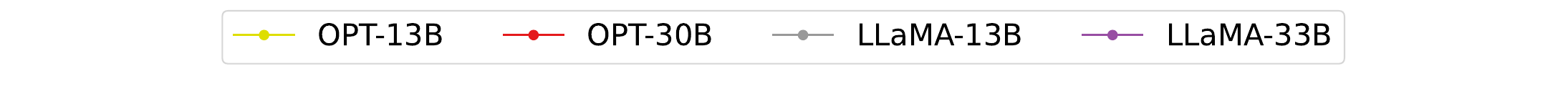}
  \includegraphics[width=1.\textwidth]{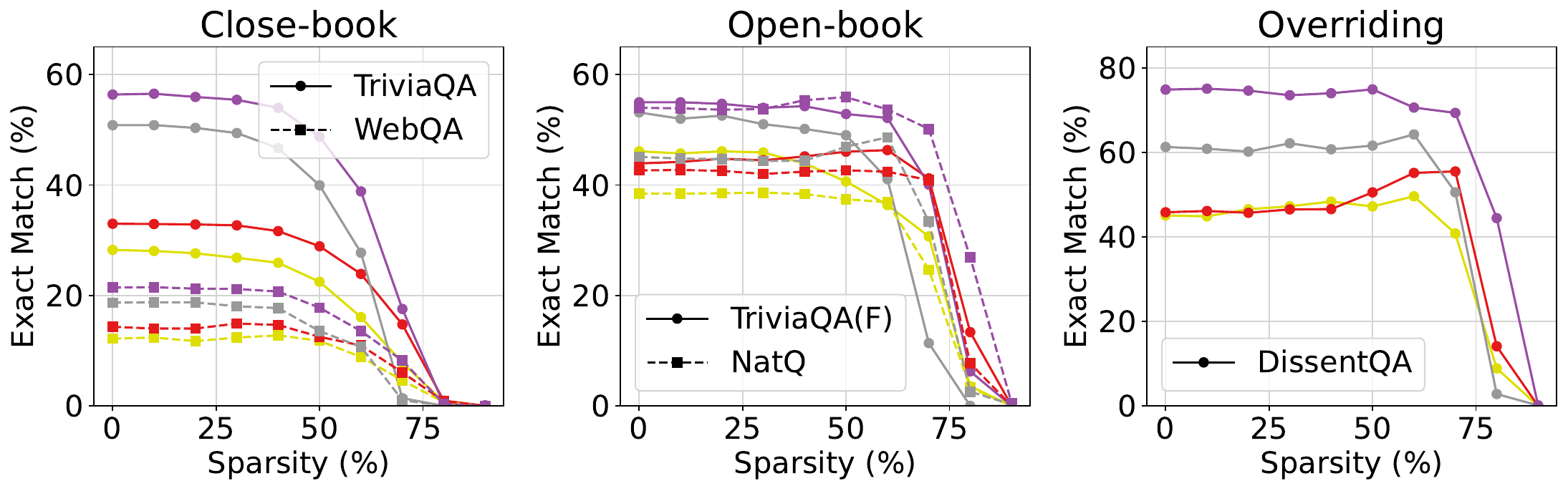}
  \caption{Evaluation for question answering tasks under pruning. 
Each color represents a different model;
each line-style/marker-shape combination represents a different QA dataset.
  Moderate pruning harms LLMs' ability to recall facts learnt during pre-training:
  in close-book QA tasks, accepting a 5\% drop in average accuracy w.r.t. the dense models, we can prune to 30 and 40\% on TriviaQA and WebQA dataset.
  However, when we present the model with the necessary information to answer the question in its context, the model's ability to answer questions survives to higher sparsity:
  on open-book QA tasks, accepting the same drop in average accuracy, we can prune to 50\% and 60\% on TriviaQA(F) and NatQ dataset.
  On the overriding task (DissentQA dataset), we may even prune to 70\% while maintaining the same acceptable relative accuracy drop.}
  \label{fig:triviaqa}
\end{figure}

\textbf{Datasets.}
We use these datasets:
(a) \emph{TriviaQA.} \citet{JoshiTriviaQA2017} developed the TriviaQA dataset with questions and supporting evidence.
We use its Wikipedia validation partition consisting of 7993 questions.
(b) \emph{WebQuestions.} \citet{berant-etal-2013-semantic} collected question-answer pairs from the Freebase knowledge database.
We use its test set consisting of 2032 questions.
(c) \emph{NaturalQuestions.} \citet{47761} compiled the NaturalQuestions dataset from Google search queries.
We sampled a 7700-question subset of its validation partition (the same size as the following dataset derived from it), to make our evaluation computationally feasible.
(d) \emph{DissentQA.} \citet{neeman2022disentqa} constructed the DissentQA dataset from the NaturalQuestions dataset.
It contains pairs of questions and evidence for a made-up answer that is different from the factual one. 
It assesses whether the model can override its memory formed during pre-training with new context.
We use its validation partition consisting of 7700 questions.

\textbf{Evaluation setup.}
Using the aforementioned datasets, we evaluate the ability of pruned models on the following task setup:
\textbf{(i) \emph{Close-book}}.
We feed the question without any supporting information to the model.
We use the Wikipedia partition of the TriviaQA dataset without the supporting evidence and the WebQA dataset for this setup.
\textbf{(ii) \emph{Open-book}}. 
We feed the question with supporting evidence to the model in its context.
It is important to note that this evaluation setup only evaluates whether the answer is right; it is however agnostic to the {\em mechanism} by which the answer is retrieved: the model may either generate its answer using the supporting evidence, or by recalling facts from pre-training. 
To create this dataset, we use a subset of the TriviaQA dataset whose context can fit within the maximum sequence length of the model, consisting of 881 questions.
We denote this filtered subset as TriviaQA(F).
Additionally, we use the NaturalQuestions dataset for this setup with factual contexts.
\textbf{(iii) \emph{Overriding}}.
We present the question accompanied by evidence that deviates from the facts presented during pre-training.
Given that the anticipated made-up answers are randomized and different from the factual ones, the model cannot depend on memorization from its pre-training data to generate responses. 
This evaluation framework rigorously assesses the model's ability to override its pre-training data with new, context-specific information.
We use the DisentQA dataset with synthetic contexts for this setup.
Examples of the setups are shown in \cref{app:example-prompt}.

To summarize: across these setups, our prompts have three parts to them: first, (1) an example context--question--answer triplet (or question--answer pair in close-book setting) for demonstration,  
(2) the supporting evidence for a test question (except in the close-book setup), and (3) a test question to answer. 
Answers are the model's prompt completions produced by greedy decoding.
We report the percentage of answers that exactly match ground truth.

\textbf{Close-book versus open-book results.} Left 2 plots of
\Cref{fig:triviaqa} shows close-book and open-book results.
Notably, for all the models, the pruned model maintains performance on open-book tasks until much higher sparsity levels compared to close-book tasks.
In particular, when we accept a relative decrease of 5\% from pruning in the mean accuracy over four models, 
the highest achievable sparsities for closed-book tasks are 30\% on TriviaQA and 40\% on WebQA. 

In contrast, maintaining the same acceptable performance drop, the highest acceptable sparsity levels for open-book tasks are 50\% on TriviaQA(F) and 60\% on NaturalQuestions.
Our results suggest that while pruning hurts the model's ability to recall information from its pre-training data at moderate sparsity (30--40\%), 
one possible remedy is to provide the model with relevant evidence in context.
The model's ability to answer questions with relevant evidence in its context remains largely intact to higher sparsity levels (50--60\%).

\textbf{Overriding results.}
The rightmost plot of \Cref{fig:triviaqa} demonstrates a consistent trend for the overriding tasks. 
On the DisentQA dataset with overriding context, the highest achievable sparsity is 70\% allowing for the same 5\% accuracy drop as before. Recall that to solve the overriding task, the model must rely
on in-context information rather than in-weights information from the pre-training data. 
Thus, our observations here further substantiate that the model's ability to extracting information from its context remains largely intact at higher sparsity levels. 

\textbf{Accuracy improvements.}
We note that, surprisingly, on open-book and overriding tasks, there can be small accuracy boosts from pruning.
For example, the pruned OPT-30B sees a 9.7\% accuracy improvement on DissentQA dataset in overriding task setup. The pruned LLaMa-13B sees 3.5\% accuracy improvement on NaturalQuestions in the open-book task setup.
Though prior work often reports accuracy improvements with pruning \citep{frankle2018the, Renda2020Comparing}, we are, to the best of our knowledge, the first to observe such improvement in the QA setup.

\textbf{Takeaways.}
Our findings suggest that scaling via
pruning has a much higher impact on the ability to retrieve facts from pre-training than on the ability to retrieve information from the context. 
Pruning may even improve model's ability to answer questions when its context includes the necessary information to answer the question. 

\section{More Sophisticated In-context Learning}
\label{sec:icl}

\Cref{sec:qatasks} demonstrates that moderate pruning preserves question answering task accuracy when relevant information is available in context.
In this section, we show that even with more sophisticated ICL tasks than previously studied, moderately pruned models maintain their ICL abilities.

Typical complex ICL tasks require extensive knowledge about language and the world gathered from pre-training.
However, as \cref{sec:qatasks} shows, pruning likely impairs such knowledge. 
So, to isolate LLMs' ICL ability from factual recall abilities, 
we choose a set of tasks where the goal is to learn a \textit{parameterized} function (e.g., a linear model) through input-output examples present in-context.

\begin{figure}
  \includegraphics[width=1.\textwidth]{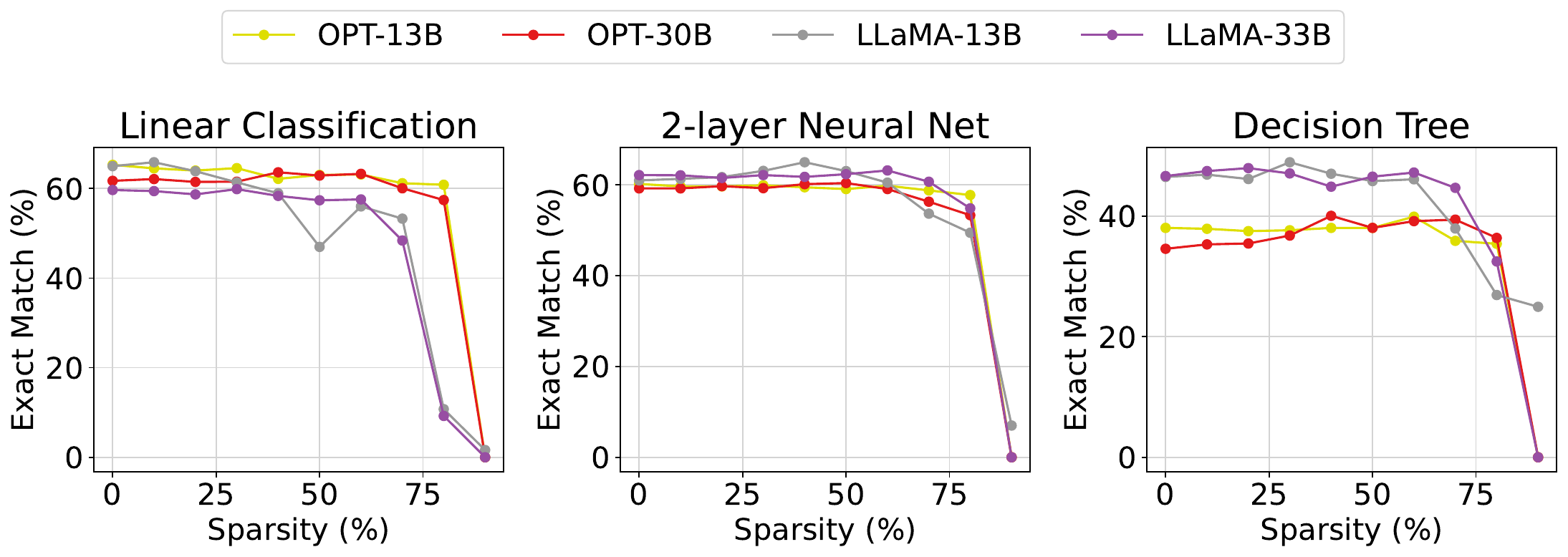}
  \caption{ICL withstands aggresive pruning (60-70\% spasity).
  Accepting a relative average (over 4 models) accuracy drop of 5\% w.r.t. the dense models, we can prune to 60\%, 60\% and 70\% on linear classification, 2-layer NN, decision tree tasks, respectively.
  }
  \label{fig:icl-linear-regression}
\end{figure}

\textbf{Method.}
We evaluate LLMs in their ability to learn multiclass classifiers of the form $f: \mathbb{R}^{D} \to \{ 0, 1, \cdots K-1 \}$ from in-context examples. 
To generate a sequence consisting of a context, query and an answer, we first pick a random parameterized function $f$ from one of three function classes: 2-way linear classifiers, 2-way neural network classifiers with 2 layers and 4-way decision tree classifiers
(\Cref{sec:constructing-f} contains details for constructing $f$).
Then, in the context, we provide $N$ exemplars of the form $(x, f(x))$, followed by a novel query $x_{\rm query}$. The goal of the model is to correctly predict the answer as $f(x_{\rm query})$. %
Please refer to \Cref{app:example-prompt} for an example of the full prompt.

All task instances consist of $N$ total in-context examples split evenly across all possible labels.
We draw all random numbers uniformly from integers between -10 and 10 inclusive.
We set 
D=4, N=32.
We record additional methodological details in \Cref{sec:icl-details}.

\textbf{Pruning results.}
\Cref{fig:icl-linear-regression} shows the accuracy of pruned models versus sparsity.
We can remove a substantial number of parameters from the models without affecting their ICL ability.
Specifically, the average accuracy of the pruned models on linear, 2-layer NN, decision tree classification tasks is within 5\% of the dense models up to 60\%, 60\% and 70\% sparsity, respectively.

\textbf{Task difficulty.}
One trivial explanation for our observation is that perhaps these tasks are so easy that they are solvable even by a severely pruned model. To eliminate this hypothesis, in \Cref{sec:increasing-icl-dim}, we increase the input dimension ($D$) of the linear classification task. Even as we increase this to a significant extent (say, to an extent that the task is almost, but not totally, unlearnable), we find that the ICL performance is maintained under aggressive pruning.

\textbf{Conclusion.}
Our results suggest that when pruning LLMs moderately (i.e., to $<70\%$ sparsity), the model's ICL ability stays largely intact.
Thus, unlike fact recall tasks,
one can prune models significantly more while maintaining ICL performance.

\section{Dense Scaling}
\label{sec:dense-scaling}

\begin{figure}[b]
    \centering
    \begin{minipage}[b]{\textwidth}
    \includegraphics[width=1.\textwidth, trim={0, 11.5cm, 0, 0}, clip]{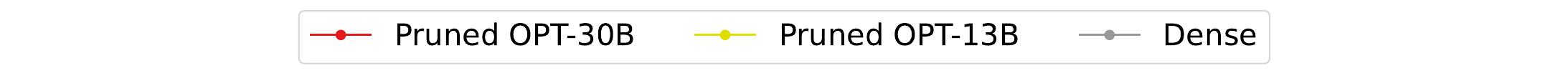}
    \end{minipage}
    \begin{minipage}[b]{0.31\textwidth}
        \includegraphics[width=\textwidth]{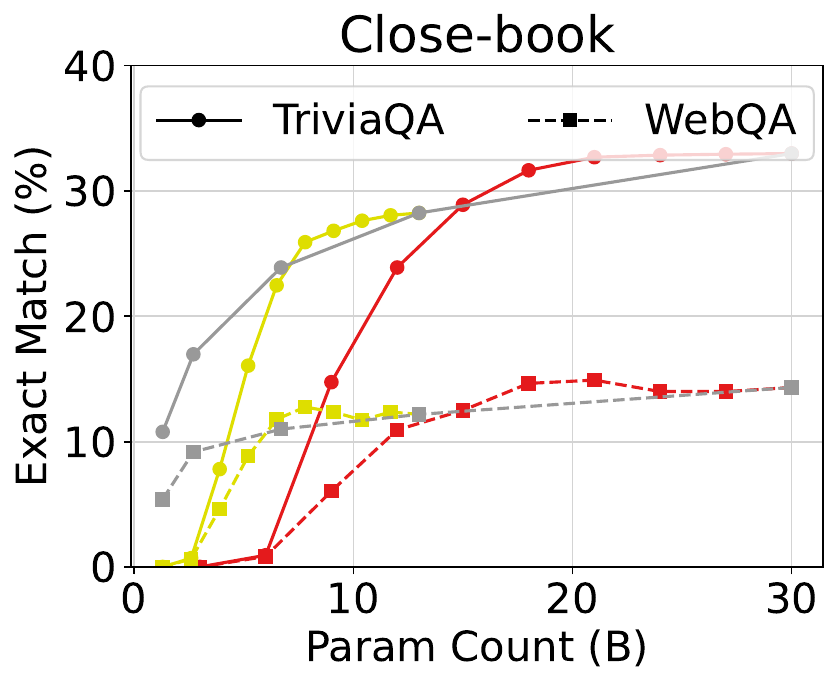}
        \label{fig:dense-scaling-qa-1}
    \end{minipage}
    \hfill %
    \begin{minipage}[b]{0.31\textwidth}
        \includegraphics[width=\textwidth]{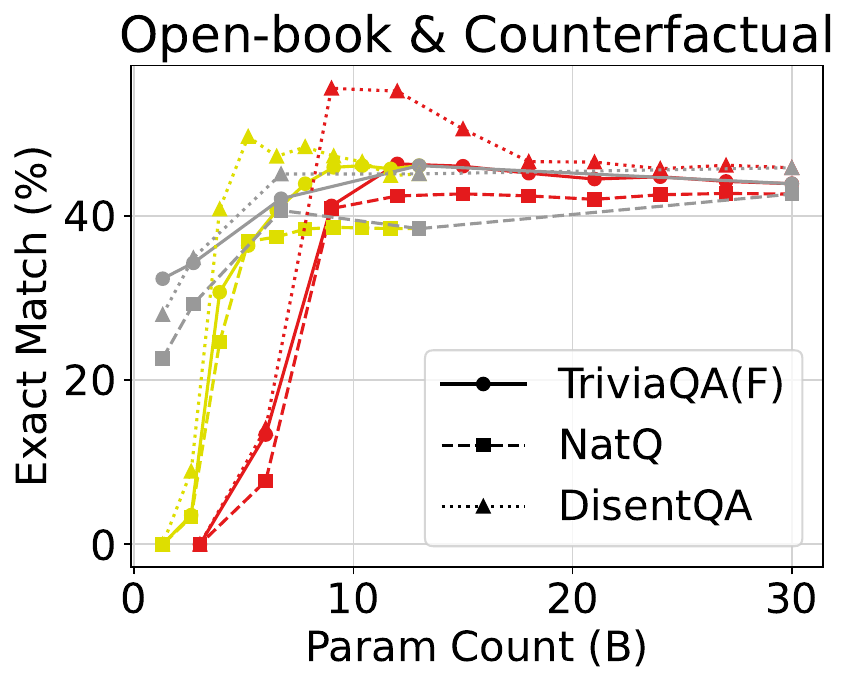}
        \label{fig:dense-scaling-qa-2}
    \end{minipage}
    \hfill
    \begin{minipage}[b]{0.30\textwidth}
        \includegraphics[width=\textwidth]{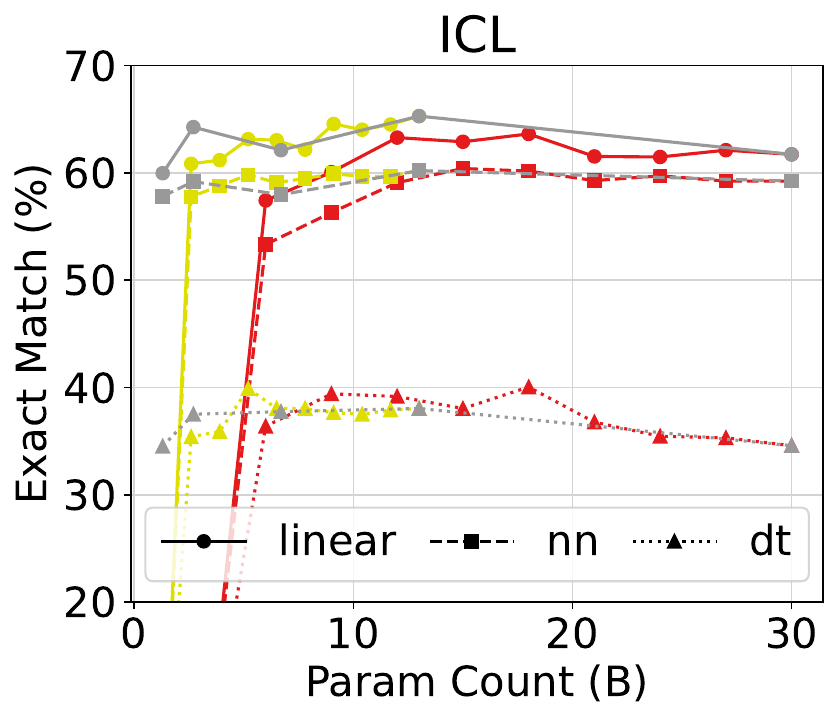}
        \label{fig:icl-sparse-vs-dense}
    \end{minipage}
    \vspace{-1em}
    \caption{Like pruning, moderate dense down-scaling preserves ICL while harming fact recall.}
    \label{fig:dense-scaling}
\end{figure}

In the context of vision models, it is known that pruned models behave similarly to smaller dense models trained with similar hyperparameters \citep{NEURIPS2022_f7ede941}.
Here, we examine the effects of dense scaling (i.e., choosing from a suite of independently trained models of various sizes) on fact recall vs. ICL abilities of LLMs.

\textbf{Method.}
With the same benchmarks as in \Cref{sec:qatasks,sec:icl}, we evaluate the accuracy of dense models with different parameter count, but the same pre-training data size.
The OPT family of models provides an ideal collection of dense models for our evaluation:
\citet{zhang2022opt} trained all OPT models with the same number of tokens (300b).

\textbf{Results.}
\Cref{fig:dense-scaling} shows that fact recall ability is highly sensitive to dense scaling -- e.g., focusing on scaling downwards, moving from the 30B model to the next largest 13B model leads to more than 5\% relative task accuracy degradation on close-book TriviaQA and WebQA task. 
However, open-book/overriding QA accuracy is much less sensitive to dense scaling.
Specifically, maintaining the same 5\% acceptable relative accuracy degradation with respect to the 30B models, one may replace a 30B model on TriviaQA(F) and NaturalQuestions dataset with a 6.7B model.
\Cref{fig:dense-scaling} also shows that ICL ability is similarly robust to dense scaling.

\textbf{Conclusion.}
Like pruning, changing a dense model's size more readily affects its ability to retrieve facts from pre-training than to process information from context.
We hypothesize that this effect stems from scaling in general -- be it pruning or using a dense model of different size.

\section{Pruning Attention and Feed Forward Layers}

\label{sec:pruning-att-vs-ffw}

A growing line of work \citep{Dai2021KnowledgeNI,NEURIPS2022_6f1d43d5} suggests that feed forward (FFW) layers and the attention layers are responsibly for distinct capabilities considered in this paper --- namely, fact recall and in-context learning.  
To test this, we exclusively prune either type of layers and examine how the model's abilities deteriorate on fact recall and ICL tasks.

\begin{wrapfigure}{r}{0.5\textwidth}
  \begin{subfigure}{0.24\textwidth}
    \includegraphics[width=\linewidth]{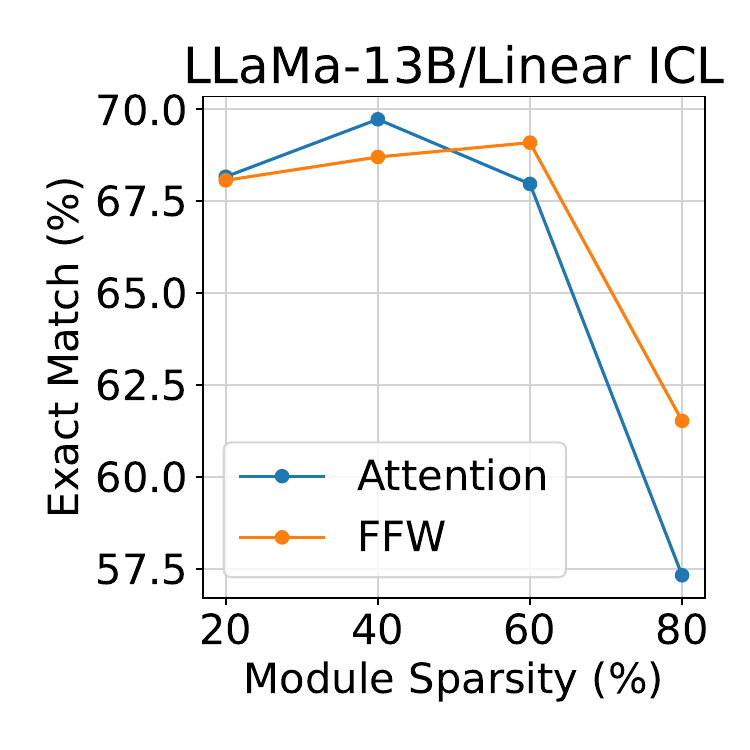}
    \caption{Linear Classification}
  \end{subfigure}
  \begin{subfigure}{0.24\textwidth}
    \includegraphics[width=\linewidth]{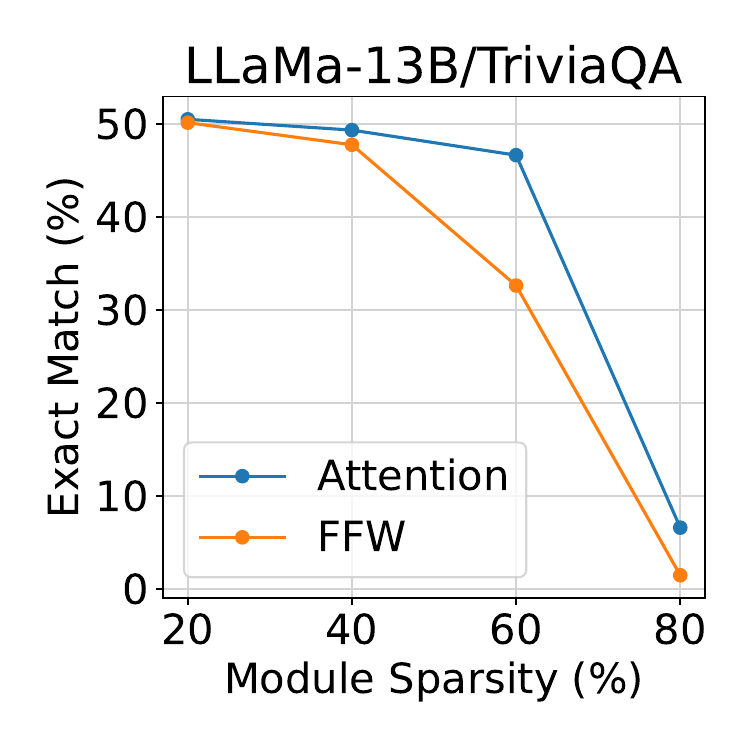}
    \caption{TriviaQA Close-book}
  \end{subfigure}
  \caption{Effects of pruning only attention layers versus pruning only FFW layers on fact recall (TriviaQA) versus ICL (Linear Classification). 
  Module sparsity computes the fraction of parameters of the specified module ( attention or FFW) that pruning removes.
    Attention and FFW layers appear equally important for ICL. FFW layers appear more important than attention layers for fact recall.}
   \label{fig:attn-vs-ffw}
\end{wrapfigure}

\textbf{Method.} We prune a LLaMA-13B model. 
We either exclusively prune attention layers or exclusively prune FFW layers and observe their effects on accuracy of TriviaQA task (fact recall) and linear classification task (ICL). 
We plotted the task accuracy as a function of module sparsity. 
Module sparsity refers to the fraction of parameters pruned from a specific type of layers (either attention or FFW), with respect to the total number of parameters of that type.

\textbf{Results.} \Cref{fig:attn-vs-ffw} shows that while attention and FFW layers appear equally important for ICL, FFW layers appear more important for fact recall. 
For example, pruning 60\% FFW layers lead to 14\% more accuracy degradation than pruning 60\% of attention layers.

\textbf{Conclusion.} Attention and FFW layers show similar importance for ICL; and FFW layers show greater importance for knowledge recall.

\section{Closing Discussion}

We study the effects of scaling model size via pruning and dense scaling on two core %
abilities of LLMs that underpin their practical success: the ability to recall facts and the ability to perform ICL. 
In both cases, we find similar disparate effects on the two abilities. %
Moderate pruning ($>30\%$ sparsity) harms fact recall, and yet the ability to learn from a few input-output examples from context withstands aggressive pruning (up to 60--70\% sparsity). The same disparity arises 
when changing the width and depth of dense (independently-trained) models. 

What could explain this disparity? 
We conjecture that the number of parameters required to store a set of facts must scale in proportion to the number of independent facts. 
On the other hand, in-context learning of many kinds may be accomplished by a smaller set of parameters that act as a universal gradient descent module \citep{vonoswald2022transformers} that can be applied for any task.  %
Verifying these hypotheses theoretically is an important direction for future work. 
Our findings also invite various research directions for practice:

\textbf{Pruning \& interpretability.}
Our findings suggest, remarkably, a relatively small fraction of weights are responsible for ICL performance.
This observation could prove to be valuable for enhancing the interpretability of LLMs, reviving a decades-old motivation behind work in pruning \citep{NIPS1988_07e1cd7d}.%
\footnote{"This skeletonization technique can be used ... to understand the behavior of networks in terms of minimal 'rules.'" \citep{NIPS1988_07e1cd7d}} 
In particular, pruning may help better localize the weights responsible for in-context ability, and complement recent approaches \citep{wang2023interpretability,bills2023language}.

\textbf{Memory augmentation.}
Our observations advocate for {\em memory augmentation} as a promising way to improve the trade-off between computational cost and task accuracy. Memory augmentation techniques present helpful facts to the model by augmenting them directly in the context. Thus, rather than having the LLM to rely on fact recall from pre-training  --- an ability that is degraded under downscaling --- we can delegate fact-retrieval to a separate retrieval model \citep{pmlr-v162-borgeaud22a,10.5555/3524938.3525306}, and allow the LLM to focus on retrieving the fact from the context---which is an ability that \textit{is} preserved under down-scaling.

\textbf{Limitations.} We address the limitations of our work in \Cref{sec:limitation}.

\bibliography{iclr2024_conference}
\bibliographystyle{iclr2024_conference}

\clearpage

\appendix

\section{Limitation.}
\label{sec:limitation}

Overall, our study provides a valuable first step towards evaluating pruning algorithms for LLMs, as well as identifying limitations of a particular type of pruning approaches. 
We have validated our claims on a large number of benchmarks (a total of 8 tasks and 6 models). 
Since our work is empirical in nature, our observations may not generalize to the full spectrum of tasks and large language models (LLMs). 
In this study, we focused on evaluating pruning algorithms that scale to large-scale LLMs. 
This was a deliberate decision, as we believe that scalability is essential for any pruning algorithm to be widely adopted in the real world.
More sophisticated one-shot/iterative pruning algorithms exist \citep{Renda2020Comparing}. 
They typically require re-training -- redoing the training of these foundation models for every sparsity level we examine.
The cost of such an experiment is at least in the millions of dollars therefore beyond our means.

\section{Pruning Perplexity Eval}
\label{sec:pruning-perplexity-eval}
In \Cref{tab:pruning-ppl} we present the full range of pruning's effect on the perplexity of models we consider in this work.
We obtain perplexity results by running the pruned model on a randomly sampled subset of C4 validation set, following the precedent of \citet{frantar2023gptq}.

\begin{table}[]
\begin{tabular}{@{}lllllllllll@{}}
\toprule
Name/Sparsity & 0.0  & 10.0 & 20.0 & 30.0 & 40.0 & 50.0 & 60.0 & 70.0 & 80.0 & 90.0    \\ \midrule
OPT-13B       & 11.5 & 11.5 & 11.6 & 11.6 & 11.9 & 12.4 & 13.7 & 17.8 & 45.3 & 40362.3 \\
OPT-30B       & 10.9 & 10.9 & 10.9 & 11.0 & 11.1 & 11.4 & 12.2 & 14.6 & 35.4 & 4187.5  \\
LLaMa-13B     & 6.6  & 6.6  & 6.7  & 6.8  & 7.1  & 7.8  & 10.0 & 20.0 & 59.9 & 822.1   \\
LLaMa-33B     & 5.9  & 6.0  & 6.0  & 6.1  & 6.4  & 7.0  & 8.5  & 13.7 & 36.1 & 355.9   \\ \bottomrule
\end{tabular}
\caption{Pruning's effect on perplexity.}
\label{tab:pruning-ppl}
\end{table}

\begin{figure}
    \centering
    \includegraphics[width=1.\textwidth]{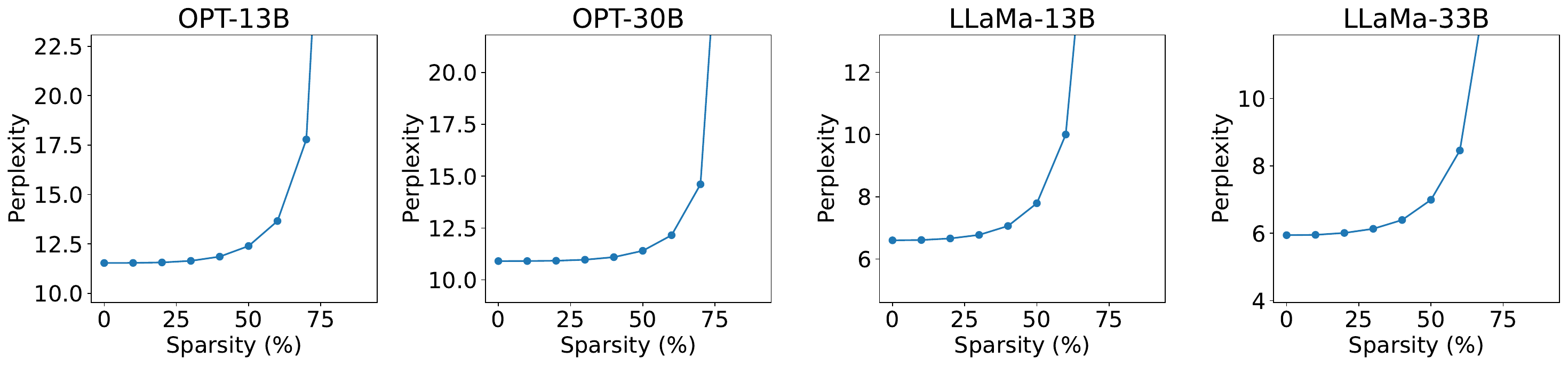}
    \caption{Pruning's effect on model perplexity.}
    \label{fig:my_label}
\end{figure}

\section{Constructing Parameterized Functions in ICL Evaluation}
\label{sec:constructing-f}
While it may be unreasonable to expect the model to infer any arbitrary $f$, we focus on three natural classes of functions from which we pick $f$: the class of linear, 2-layered neural network, and decision tree models. These classes are defined as follows:
(a) \emph{Linear.}
For each task instance, we generate a distinct random hyperplane in the $D$-dimensional input space as the decision boundary.
We label each example as positive/negative depending on the side of the decision boundary it falls on.
(b) \emph{2-layer NN.}
For each task instance, we generate a 2-layer neural network mapping a $D$-dimensional input vector $x$ to a binary label $y$ with the following form: $W_2\sigma(W_1x)$, where $W_1$, $W_2$ are $D \times D$, $D \times 2$ matrices correspondingly.
We draw elements within these two matrices from independent, Gaussian distributions with zero-mean and unit variance.
We use ReLU as the activation function $\sigma$.
(c) \emph{Decision tree.}
For each task instance, we construct a full depth-2 binary decision tree, with each of its four leaf nodes representing one possible label. 
This maps $D$-dimensional input vectors to the said labels. 
We assign each non-leaf node an index from 0 to $D-1$. 
We evaluate the tree by traversing from the root, comparing the input vector's value at each node's associated index to determine the next step: if negative, we go left; if zero or positive, we go right. 
Reaching a leaf node marks the end of traversal, and the corresponding label represents the evaluation result.

\section{Additional Details for ICL Evaluation Methodology}
\label{sec:icl-details}

In this section, we provide additional details for our ICL task evaluation methodology.

\paragraph{In-context example label distribution.}
When generating the ICL evaluation datasets, we programmatically enforce that the labels for the $N$ in-context demonstrations are split exactly evenly between K possible labels.

\paragraph{Evaluation label distribution.}
Each evaluation input has an equal probability of receiving one of the $K$ possible classification labels.
Therefore, evaluation label distribution should be even in expectation.
However, we do not programmatically enforce an even distribution of evaluation labels.

\paragraph{Size of evaluation dataset.}
We generate 2048 independent task instances for each function class.
Each task instance represents a unique parameterization of the classification function to learn.

\paragraph{Prompting technique.}
We do not use any advanced prompting technique when solving ICL tasks.
The LLMs only generate answer tokens corresponding directly to classification labels.
We do not generate any other tokens during evaluation.

\paragraph{Tokenizer.}
The OPT models use a byte-pair encoding tokenizer whereas the LLaMA models use a sentence piece tokenizer.
Notably, LLaMA tokenizers split each digit into an individual token.
This feature of LLaMA tokenizers boosts LLaMA performance on arithmetic tasks \citep{liu2023goat}.

\section{Increasing Input Dimension of ICL Evaluation}
\label{sec:increasing-icl-dim}

In this section, we further evaluate the effect of pruning on ICL tasks with different input dimensions.

\paragraph{Method.}
In \Cref{sec:icl}, we evaluate pruning's effect on learning linear classification tasks in-context.
The task is to classify 4-dimensional input vectors.
Here, we increase the input vector dimensions to 8 and 16, and observe whether our observation about pruning's effect generalize to larger input dimenions.

\begin{figure}[h!]
    \centering
    \includegraphics[width=1.\textwidth]{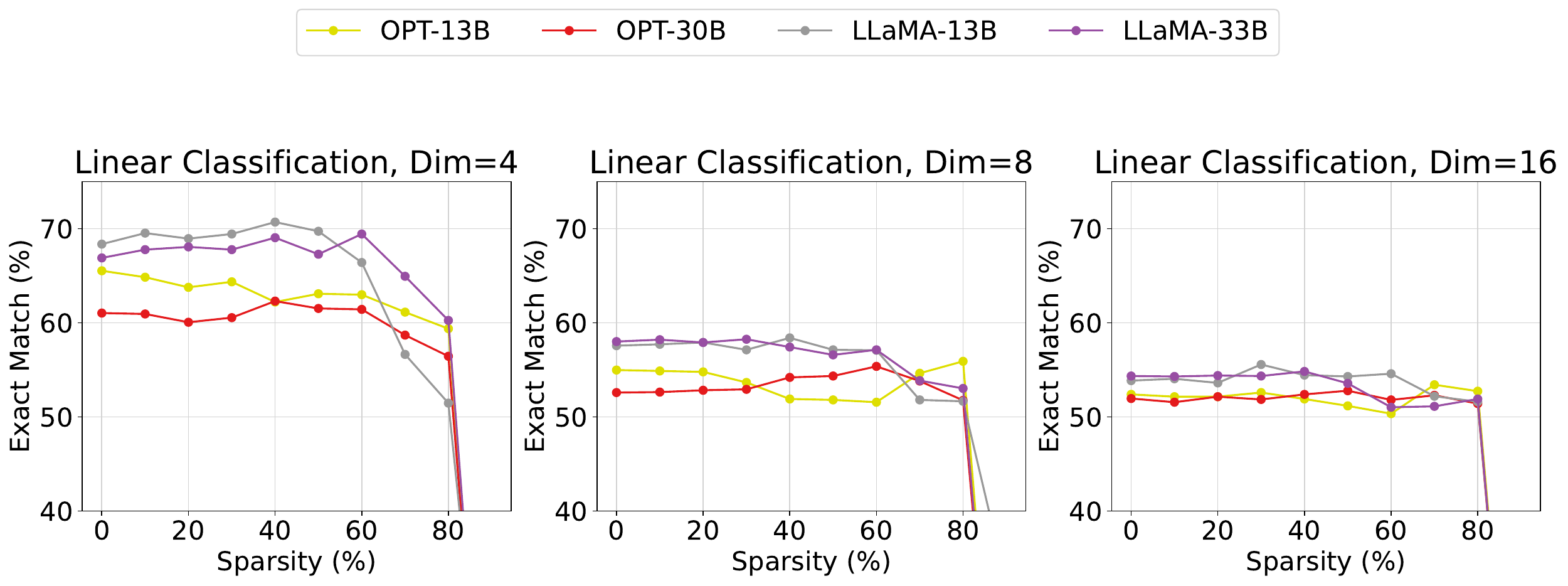}
    \caption{Pruning's effect on ICL linear classification tasks with input dimension of 4, 8 and 16.
             We observe that even with increased input dimension, ICL ability of LLMs continue to be robust to aggresive pruning.}
    \label{fig:icl-higher-dims}
\end{figure}

\paragraph{Results.}
\Cref{fig:icl-higher-dims} presents the relationship between task accuracy and sparsity for linear classification task.
We increase the input dimension from 4 to 8 and 16.
We observe that tolerating the same 5\% relative drop in the average accuracy among all four models with respect to the dense model, we can prune to 60\%, 80\% sparsity on linear classification task with 4, 8 dimensional inputs respectively.
Dense models perform close to chance on 16-dimensional inputs, thus we refrain from drawing conclusions based on these results.

\paragraph{Conclusion.}
Even with higher dimensional inputs, model's ICL ability remains resilient to aggressive pruning ($>=60\%$ sparsity).

\section{Additional Pruning Algorithm Evaluation}
\label{sec:another-pruning-algo}
In our main paper, we focus on a single pruning algorithm SparseGPT \citep{frantar2023sparsegpt}.
In this section, we study whether our observation that scaling down LLM size affects fact recall more readily than ICL generalize to another pruning algorithm.

\paragraph{Method.} We repeat key experiments with another pruning algorithm called Wanda \citep{sun2023simple}.
Notably, unlike SparseGPT, Wanda does not update the remaining weights after weights removal. The author of Wanda shows that at 50\% sparsity, Wanda achieves accuracy that is competitive with SparseGPT.

\paragraph{Results.} We observe the same disparate effects from pruning on fact recall versus ICL: while moderate pruning hurts fact recall, ICL survives to higher sparsity. Specifically, accepting the same 5\% relative drop in accuracy as we did for SparseGPT results, one may remove 30\%, 40\% and 50\% weights on TriviaQA(Closebook), TriviaQA(Openbook) and Linear Classification tasks. Unsurprisingly, given that Wanda is computationally less expensive, it underperforms SparseGPT at high sparsities on ICL tasks.

\paragraph{Conclusion.} With our experiments repeated with another pruning algorithm, we show that our observation is general.

\begin{figure}
\begin{subfigure}{.33\textwidth}
    \centering
    \includegraphics[width=\linewidth]{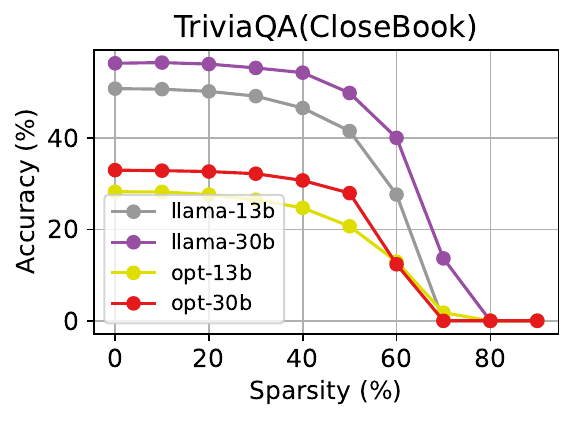}
  \end{subfigure}
  \begin{subfigure}{.33\textwidth}
    \centering
    \includegraphics[width=\linewidth]{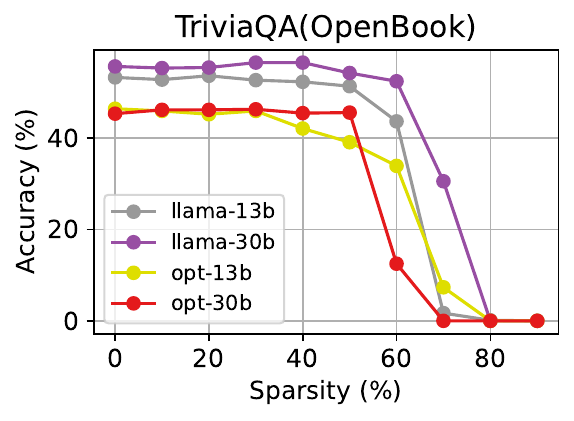}
  \end{subfigure}
  \begin{subfigure}{.33\textwidth}
    \centering
    \includegraphics[width=\linewidth]{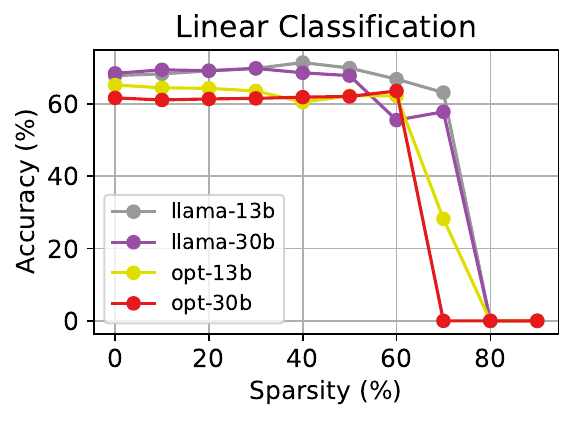}
  \end{subfigure}%
  \caption{An alternative pruning algorithm (Wanda) shows the same patterns of accuracy drop as findings in our original paper: moderate pruning hurts fact recall (e.g., left, TriviaQA in Closebook setting) while ICL (e.g., right, in-context linear classification) survives to higher sparsity. 
  Specifically, accepting a 5\% relative drop in accuracy, one may remove 30\%, 40\% and 50\% weights on TriviaQA(Closebook), TriviaQA(Openbook) and Linear Classification tasks, respectively.}
\end{figure}

\section{Pythia Models Evaluation}
\label{sec:pythia}

\begin{figure*}[t]
	\begin{subfigure}{.32\textwidth}
    \centering
    \includegraphics[width=\linewidth]{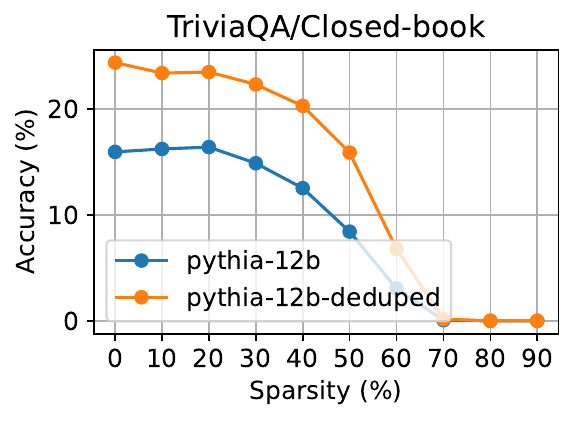}
  \end{subfigure}
  \begin{subfigure}{.32\textwidth}
    \centering
    \includegraphics[width=\linewidth]{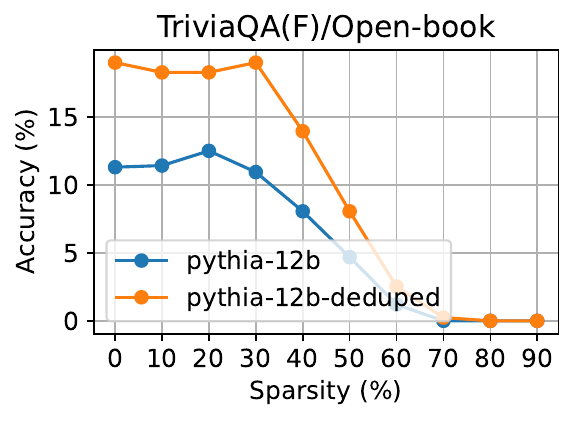}
  \end{subfigure}
  \begin{subfigure}{.32\textwidth}
    \centering
    \includegraphics[width=\linewidth]{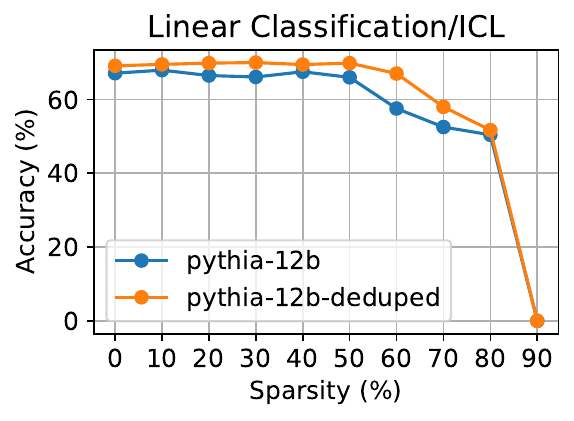}
  \end{subfigure}%
  \caption{Two variants of Pythia-12B model shows the same patterns of accuracy drop as findings in our original paper: moderate pruning hurts fact recall (e.g., left, TriviaQA in Closebook setting) while ICL (e.g., right, in-context linear classification) survives to higher sparsity. 
  Specifically, accepting a 5\% relative drop in average accuracy, one may remove 20\%, 30\% and 50\% weights on TriviaQA(Closebook), TriviaQA(Openbook) and Linear Classification tasks, respectively.}
\end{figure*}

In this section, we study whether our observation that scaling down LLM size affects fact recall more readily than ICL generalize to another model family: Pythia \citep{biderman2023pythia}.

\paragraph{Method.} We repeat key experiments with Pythia model family. 
With SparseGPT, we pruned two variants of the Pythia-12B model, the original and a “deduped” variant where the model is trained on a deduplicated dataset. They're the most capable models in the Pythia family.

\paragraph{Results.}
\Cref{fig:pythia} shows consistent disparate effect of pruning on fact recall versus ICL. 
Specifically, accepting the same 5\% accuracy drop as in all our results, one can remove 20\%, 30\% and 50\% weights for TriviaQA(Closebook), TriviaQA(Openbook) and in-context linear classification tasks.

\paragraph{Conclusion.} Our observation is robust to the choice of model families.

\section{Example Prompts}
\label{app:example-prompt}

In this section, we present example prompts for Q\&A tasks in \Cref{sec:qatasks} and ICL on algorithmic tasks in \Cref{sec:icl}.

\subsection{Question-answering without context.}
Our prompt follows the precedent of \citet{touvron2023llama}.
We show our prompt for Q/A tasks without context in \Cref{prompt:qa-without-context}.

\begin{figure}
\fbox{
\begin{minipage}{35em}
Answer these questions:

Q: In Scotland a bothy/bothie is a?

A: House

Q: Which 90s sci fi series with James Belushi was based on Bruce Wagner's comic strip of the same name?

A:
  \end{minipage}}
  \caption{Examples Prompt for Question-answering Task without Context.}
  \label{prompt:qa-without-context}
\end{figure}

\begin{figure}
\fbox{
\begin{minipage}{35em}
Answer these questions:

Context: Scattered across the Highlands and rural areas of Scotland, a bothy is a small house that can be used by anyone.

Q: In Scotland a bothy/bothie is a?

A: House

Context: "Roses Are Red (My Love)" is a popular song composed by Al Byron and Paul Evans. It was recorded by Bobby Vinton and was his first hit.

The song was released in April 1962.  It reached No. 1 in Australia, New Zealand, Norway, the Philippines, South Africa, and the United States, and was a major hit in many other countries as well. The song topped the Billboard Hot 100 singles chart on July 14, 1962, and remained there for four weeks.   The single was also the first number-one hit for Epic Records. Billboard ranked the record as the No. 4 song of 1962.

Vinton found the song in a reject pile at Epic Records. He first recorded it as an R\&B number, but was allowed to re-record it in a slower more dramatic arrangement, with strings and a vocal choir added.

Ronnie Carroll version

In the UK, a cover version by Ronnie Carroll reached No. 3 on the Record Retailer chart on August 8, 1962, the same week that the Bobby Vinton record peaked at No. 15.   It peaked at No. 7 in the very first Irish Singles Chart published in October 1962.

Other versions

The song was recorded by Jim Reeves in 1963 and released on the album Gentleman Jim, one of the last albums released while he was still alive. While it did not chart in the US, it became a minor hit in Norway and Germany.

The song was covered by Singaporean female artist Zhuang Xue Fang, in edited Standard Chinese lyrics written by Suyin under title name of, with Ruby Records in 1967.

In 1962, an answer song, entitled "Long As The Rose Is Red", was recorded by Florraine Darlin.  The song spent seven weeks on the Billboard Hot 100, reaching No. 62,  while reaching No. 15 on Billboards Easy Listening chart.   It was released by Epic Records (single \#9529) and was also produced by Robert Morgan.

Charts

Bobby Vinton version

Ronnie Carroll version

Q: Which singer had a big 60s No 1 with Roses Are Red?

A:
\end{minipage}}
  \caption{Examples Prompt for Question-answering Task with Context.}
  \label{prompt:qa-with-context}
\end{figure}

\subsection{Question-answering with context.}
We show an example prompt we use to evaluate Q/A task with the necessary information to answer the question available in-context in \Cref{prompt:qa-with-context}.

\subsection{ICL on algorithmic task.}
We present the prompt we use to evaluate ICL on linear classification task in \Cref{prompt:linear-classification}.
Evaluations for ICL on check-sorting, check-even  uses analogous prompts.

\begin{figure}
\fbox{
\begin{minipage}{35em}
[5, -5, -1, 6] = -1 

[3, 5, -9, -2] = 1 

[-10, 5, -9, -10] = 1 

[7, 0, -4, -7] = -1

[-2, -2, 9, -3] = -1

[9, -5, -5, -1] = -1

[-4, -1, -10, -8] = 1

[-10, -1, -3, 6] = -1

[3, 6, -5, -8] = 1

[-1, 7, -2, 6] = -1

[-5, 8, -1, -9] = 1

[-6, -1, -6, -9] = 1

[1, 0, -10, -9] = 1

[-5, 5, -3, 6] = -1

[-4, 6, -10, -7] = 1

[-4, 5, -6, -10] = 1

[-10, 5, -3, -9] = 1

[-7, -2, -7, -9] = 1

[-7, 1, -7, 4] = -1

[-2, 2, -3, -10] = 1

[1, -4, -4, 9] = -1

[7, 0, 6, -10] = -1

[2, 7, -6, -1] = 1

[1, -2, -10, -9] = 1

[-8, 3, 7, -7] = -1

[0, 7, -3, -8] = 1

[0, 2, 0, 0] = -1

[-7, 9, 2, 8] = -1

[1, -8, -9, 6] = -1

[-5, 1, -6, -10] = 1

[8, 8, 1, -4] = -1

[-3, 6, 2, 2] = -1

[-3, 9, 0, -5] =
  \end{minipage}}
  \caption{Examples Prompt for ICL on Linear Classification Task.}
  \label{prompt:linear-classification}
\end{figure}

\clearpage

\section{Analysis on Accuracy Improvement}
On many tasks, such as ICL on algorithmic tasks in \Cref{sec:icl}, we observe accuracy improvement from pruning.
In this section, we provide further analysis on this phenomenon. 

\paragraph{Method.}
We focus on the ICL task of check-sorting and check-even from \Cref{sec:icl}, where the pruned LLaMA models see significant accuracy improvement.
We analyze the prediction of the pruned model, in relation to the following two answers:
1). \emph{Answer without context.} 
We compute the classification output with the highest log likelihood, without all the in-context examples.
The prompt thus looks like "[1, 0, 3, -5]=" without any other information such as in-context training examples.
The LLM thus classifies this input array exclusively based on likelihood of a classification label as a continuation of the prompt.
We denote this classification outcome as the answer without context.
2). \emph{Ground-truth answer.}

\begin{figure}[h!]
    \centering
    \includegraphics[width=1.\textwidth]{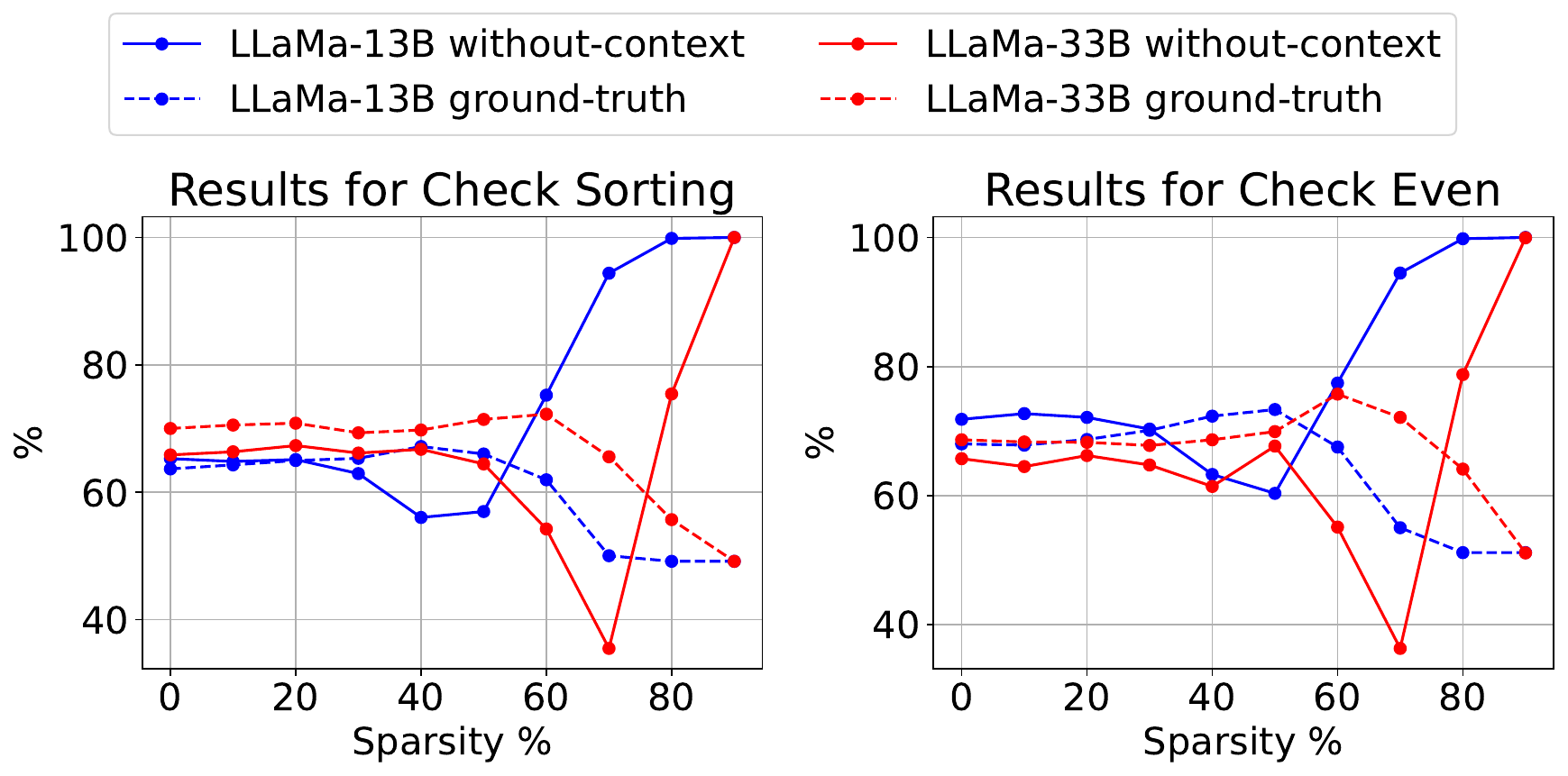}
    \caption{Analyzing the accuracy improvement of pruned model on ICL tasks.}
    \label{fig:accuracy-improvement}
\end{figure}

\paragraph{Results.}
In \Cref{fig:accuracy-improvement}, we present the percentage of time when the prediction of the pruned model matches the answer without context versus the ground truth.
We observe that accuracy improvement often mirrors a departure of the model's prediction from answer without context (the solid line dropping whilst the dashed line raising).
This suggest that for some pruned models (for example LLaMA-30B at 60\% sparsity), the presence of in-context examples changes the prediction of the LLM more than the dense models.

\paragraph{Conclusion.}
We hypothesize, without testing, that pruning may improve task accuracy by enhancing the effect of contextual information on its prediction.

\clearpage

\section{Details on Computing Software/Hardware}
\label{sec:hwsw}
We perform pruning on 80GB version of A100 GPUs, using the implementation of SparseGPT\citep{frantar2023sparsegpt}.
We perform our evaluations using TPU v3 running PyTorch \citep{NEURIPS2019_9015} with XLA backend.
Our research project consumes approximately 3 TPU-month worth of computation.

Our computing software/hardware necessitates the use of the bfloat16 numerical format. 
We assess its impact on our experimental results in \Cref{sec:numerical}.

\section{Characterizing Errors from Numerical Precision}
\label{sec:numerical}
All our experiments run on TPU v3.
The TPU hardware does not support float16 (IEEE 754 half precision floating point format), which are the formats used for training the OPT/LLaMA models.
We thus use a closely related floating point format called bfloat16 (Brain Float) instead to evaluate these models.
Both bfloat 16 and float16 are floating point numbers requiring 16 bits to represent, the key difference is that bfloat16 format can represent numbers in a slightly wider range of values, but with slightly reduced precision compared with float16.
For this reason, we expect our results to contain small systematic errors.
In this section, we characterize the difference between our results in bfloat16 and standard results in float16.

\begin{figure}[h!]
    \centering
    \includegraphics[width=1.\textwidth]{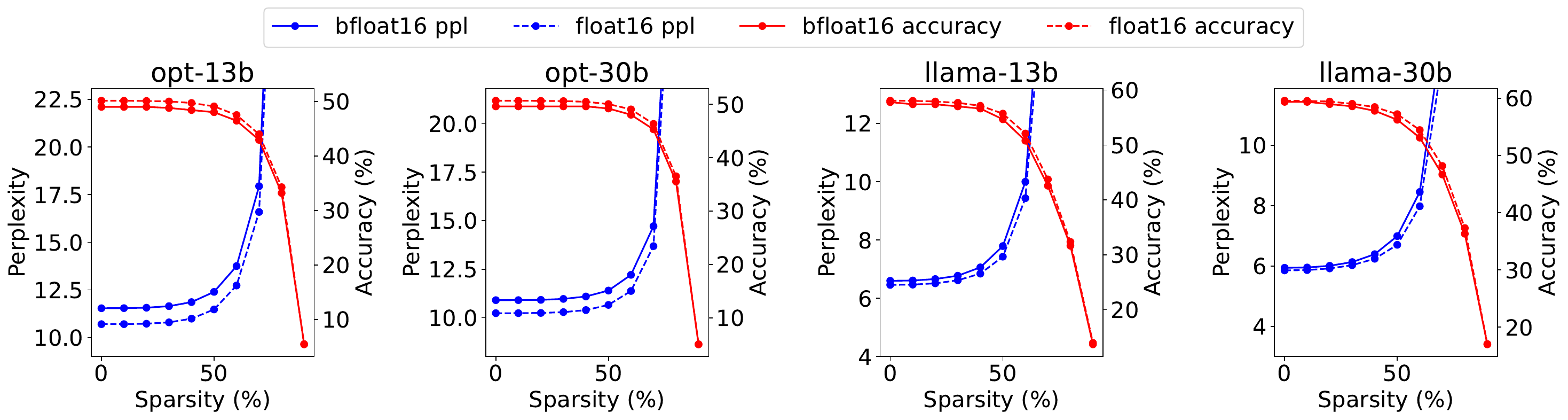}
    \caption{Characterizing the difference between float16 and bfloat16.}
    \label{fig:numerics}
\end{figure}

\paragraph{Perplexity and accuracy results.}
In \Cref{fig:numerics}, we plot the perplexity and accuracy in next token prediction in bfloat16 on TPU v3 and float16 on A100.
We observe that the average accuracy difference is 1.0\%, 0.89\%, 0.70\% and 0.68\% for OPT-13B, OPT-30B, LLaMA-13B and LLaMA-33B.
The perplexity grows to very large numbers at high sparsity, so we report the average perplexity difference for sparsity level less than or equal to 50\%.
The average perplexity difference is: 0.86, 0.69, 0.19, 0.13.
We note that 1). the difference in accuracy and perplexity due to numerical precision is systematic. 
The accuracy of next token prediction raises by a similar amount across all sparsity levels. 
2). the difference in next token prediction accuracy is small (<1\%).

\paragraph{OPT.}
OPT models see higher difference in accuracy and perplexity.
This is because its GPU implementation uses mixed precision arithmetic, where the attention score computes in float32.
Our compiler and runtime, however, does not support mixed precision arithmetic, we therefore compute entirely within bfloat16.

\end{document}